\documentclass[twoside]{article}

\usepackage{answers}
\usepackage{setspace}
\usepackage{graphicx}
\usepackage{enumitem}
\usepackage{multicol}
\usepackage{mathrsfs}
\usepackage{amsmath,amsthm,amssymb}
\usepackage{algorithm}
\usepackage[noend]{algpseudocode}
\usepackage{color}
\usepackage{dsfont}
\usepackage{makecell}
 \usepackage[accepted]{aistats2020}
\usepackage[round]{natbib}

\usepackage{amsfonts, graphicx, caption, subcaption, xcolor}
\usepackage{sidecap}
\usepackage{siunitx}

\DeclareMathOperator*{\argmax}{arg\,max}

\newcommand{\E}{\mathbb{E}}

\newcommand{\R}{\mathbb{R}}

\newcommand{\vx}{\boldsymbol{x}}
\newcommand{\vtheta}{\boldsymbol{\theta}}
\newcommand{\vw}{\boldsymbol{w}}

\newcommand{\vK}{\boldsymbol{K}}
\newcommand{\vy}{\boldsymbol{y}}

\newcommand{\vk}{\boldsymbol{k}}

\newcommand{\vdelta}{\boldsymbol{\delta}}

\newcommand{\Nc}{\mathcal{N}}

\newcommand{\Xc}{\mathcal{X}}
\newcommand{\Pc}{\mathcal{P}}
\newcommand{\Qc}{\mathcal{Q}}
\newcommand{\Hc}{\mathcal{H}}

\newcommand{\ucb}{\mathrm{ucb}}
\newcommand{\lcb}{\mathrm{lcb}}
\newcommand{\oucb}{\overline{\mathrm{ucb}}}
\newcommand{\olcb}{\overline{\mathrm{lcb}}}

\usepackage{xspace}
\newcommand{\stableopt}{\textsc{StableOpt}\xspace}
\newcommand{\ouralg}{\textsc{GP-MRO}\xspace}

\newcommand{\secvar}{\vtheta}
\newcommand{\secset}{\Theta}

\newtheorem{theorem}{Theorem}

\newtheorem{lemma}[theorem]{Lemma}
\newtheorem{corollary}[theorem]{Corollary}

\begin{document}

%

%

\twocolumn[

\aistatstitle{Mixed Strategies for Robust Optimization of Unknown Objectives}

\aistatsauthor{Pier Giuseppe Sessa \And Ilija Bogunovic \And  Maryam Kamgarpour \And Andreas Krause}

\aistatsaddress{ ETH Z\"urich \And  ETH Z\"urich \And ETH Z\"urich \And ETH Z\"urich } ]

\begin{abstract}
We consider robust optimization problems, where the goal is to optimize an \emph{unknown} objective function against the worst-case realization of an uncertain parameter. For this setting, we design a novel sample-efficient algorithm \ouralg, which sequentially learns about the unknown objective from noisy point evaluations. \ouralg seeks to discover a robust and randomized \emph{mixed strategy}, that maximizes the worst-case expected objective value.
To achieve this, it combines techniques from online learning with nonparametric confidence bounds from Gaussian processes. Our theoretical results characterize the number of samples required by \ouralg to discover a robust near-optimal mixed strategy for different GP kernels of interest. We experimentally demonstrate the performance of our algorithm on synthetic datasets and on human-assisted trajectory planning tasks for autonomous vehicles. In our simulations, we show that robust deterministic strategies can be overly conservative, while the mixed strategies found by \ouralg significantly improve the overall performance.\looseness=-1
\end{abstract}

\section{Introduction}

Many real-world problems require taking decisions under uncertainty. Latter can manifest itself in the form of uncertain parameters, perturbations, or an adversary that can corrupt the decision \citep{bertsimas2011theory}.
In such problems, one often seeks to optimize an objective function while being \emph{robust} to the worst possible uncertainty realization. This can be achieved by phrasing such problems in the framework of Robust Optimization (RO) \citep{ben2009robust}. RO has found successful applications in various domains including supply chain management \citep{bertsimas2004supplychain}, portfolio optimization \citep{bental2000portfolio},  influence maximization \citep{he2016robust}, and robotics \citep{bojorgensen2018}, to name a few. 

In various practical problems, however, the objective function to be optimized is a-priori \emph{unknown}, and one can only learn about it from \emph{sequential} and \emph{noisy} point evaluations. Gaussian process (GP) optimization is an established framework for model-based sequential optimization of such unknown functions~\citep{srinivas2009gaussian}. An array of algorithms that use Bayesian non-parametric GP models \citep{rasmussen2006gaussian}, and balance exploration (learning the function globally) and exploitation (maximizing the function) have been developed over the years, e.g.,~\citep{srinivas2009gaussian, bogunovic2016truncated, chowdhury17kernelized, wang2017max, frazier2018}.

In this paper, we study the \emph{robust} optimization problem where (i) the objective function is \emph{unknown} and (ii) the goal is to be robust against the worst possible realization of its \emph{uncertain parameter}. This problem differs from the classical RO formulation where the objective function is assumed to be known, and is also different from the standard GP optimization where robustness requirement is typically not pursued.

Instead of finding a robust deterministic solution to this problem (as in \citep{bogunovic2018adversarially}), we seek to discover a randomized, i.e., \emph{mixed} strategy, from a relatively small number of noisy function evaluations. 
The primary motivation for seeking such strategies is that, in general, they can provide arbitrarily better worst-case expected performance than deterministic ones~\citep{krause2011robust, vorobeychik2014,sinha2018security}, i.e., randomization prevents a potential adversary to know the actual decision until it is realized.
Consequently, we design and use a novel GP-based \emph{sample efficient} algorithm to discover near-optimal mixed strategies. We empirically demonstrate the effectiveness of the identified robust mixed strategies in a trajectory planning task for autonomous vehicles, where deterministic strategies are shown to be overly conservative. \looseness=-1

\textbf{Related work.} Over the past couple of years, robust optimization has been extensively studied in the machine learning community. While most of the works focus on convex settings (e.g., \citep{shalev2016,namkoong2016}), more recent works also consider general non-convex objectives, e.g.,~\citep{chen2017robust,sinha2017certifying,staib2018distributionally}. Among those, \citet{chen2017robust} provide robust algorithmic strategies that are shown to be successful in several learning tasks. The proposed algorithm is based on the idea of simulating a zero-sum game between a learner and an adversary. Similar strategies have been also explored in other adversarial settings, e.g., in submodular optimization~\citep{krause2011robust,kawase2019}. Our approach is based on the similar algorithmic idea of \citet{chen2017robust}, but unlike this and other works mentioned above that assume the objective function is perfectly \emph{known} (or a maximization oracle is available), it also requires performing a non-trivial function estimation. \looseness=-1  

In non-robust GP optimization, various optimization algorithms~\citep{srinivas2009gaussian,chowdhury17kernelized,bogunovic2016truncated, contal2013parallel, wang2017max} have been proposed to sequentially optimize the unknown function from noisy and zeroth-order observations. Similarly to these algorithms, our algorithm relies on a non-parametric GP model to obtain shrinking confidence bounds around the unknown objective function. Besides the standard problem, GP optimization has been considered in several other practical settings such as contextual~\citep{krause2011contextual}, time-varying~\citep{bogunovic2016time}, safe exploration~\citep{sui2015safe}, etc. 

Recently, a novel algorithm for \emph{robust} GP optimization \textsc{StableOpt} has been proposed by \citet{bogunovic2018adversarially}. \textsc{StableOpt} discovers a \emph{deterministic} solution that is robust with respect to the worst-case realization of the uncertain parameter. This work is closest to ours, but instead of seeking deterministic solutions, our focus is on the \emph{mixed strategies} which are preferable in certain scenarios (see Section~\ref{sec:autonomous_driving}), where deterministic solutions turn out to be overly conservative. We also note that other forms of robustness have been studied in GP optimization. For instance, \citet{nogueira2016unscented,oliveira2019} consider robustness against uncertain inputs (typical in robotics applications), \citet{sessa2019noregret} study robust aspects in multi-agent unknown repeated games, \citet{williams2000,tesch2011} deal with uncontrolled environmental variables, while robustness with respect to outliers is addressed by~\citet{martinez2018practical}.

\textbf{Contributions.} We consider robust optimization of \emph{unknown} and generally non-convex objectives.
\vspace{-1em}
\begin{itemize}[leftmargin = 1em]
  \setlength\itemsep{-0.3em}
\item We propose an algorithm, \ouralg, which returns a \emph{mixed strategy}, i.e., a probability distribution over actions, that is robust against the worst-case realization of the \emph{uncertain} parameter. 
\item  Our theoretical analysis shows the number of samples required for \ouralg to discover a near-optimal robust mixed strategy. \looseness=-1
\item  We propose a variant of \ouralg which can effectively trade-off worst-case and average-case performance.
\item Finally, we consider the problem of trajectory planning in autonomous driving guided by user's evaluations. In our experiments, we demonstrate the effectiveness of the robust mixed strategies discovered by \ouralg in comparison to those identified by existing robust methods. 
\end{itemize}
\vspace{-1.2em}
\section{Problem Formulation}
\label{sec:problem_formulation}
\vspace{-0.6em}
\setlength{\belowdisplayskip}{5pt}
\setlength{\belowdisplayshortskip}{5pt}
\setlength{\abovedisplayskip}{5pt}
\setlength{\abovedisplayshortskip}{5pt}

Let $f : \Xc \times \secset \rightarrow [0,1]$ be a reward function over domain $D = \Xc \times \secset$, where $\Xc$ is a continuous and compact decision set and $\secset = \lbrace \secvar_1, \ldots, \secvar_m \rbrace$ is a finite set of parameter values. The reward function is \emph{unknown}, and we learn about it from sequential \emph{noisy} point observations, i.e., so-called \emph{bandit} feedback. At each time step $t$, we choose $\vx_t \in \Xc$ and $\secvar_t \in \secset$, and observe a noisy sample $y_t = f(\vx_t, \secvar_t) + \xi_t$, where $\xi_t\sim \Nc(0, \sigma^2)$, and $\xi_t$'s are independent over time (our approach allows also for sub-Gaussian noise). 
 
After $T$ rounds (i.e., $T$ samples), our goal is to report a strategy for selecting points in $\Xc$ that is robust against the worst-possible parameter value from~$\Theta$. 
We assume that during the optimization phase (i.e., training/simulation) one can choose $\vtheta$, while later, during the implementation (i.e., test) phase, the parameter $\vtheta$ becomes uncontrollable. Hence, it is important to design a robust strategy for selecting the first parameter.

\vspace{-0.2em}
\textbf{Optimization goal.} Let $\Delta{(\Xc)}$ denote the set of all probability distributions, or \emph{mixed strategies} on $\Xc$. Our goal is to find a distribution in $\Delta{(\Xc)}$ that achieves high reward in the worst-case over $\secvar \in \secset$. The \emph{maximin} optimal value is given by:
\begin{equation} \label{eq:maximin}
	\tau^* = \max_{\Pc \in \Delta{(\Xc)}} \min_{\secvar \in \secset} \E_{\vx \sim \Pc}[f(\vx, \secvar)],
	\vspace{-0.2em}
\end{equation}
and we seek to report a robust solution $\Pc^{(T)} \in \Delta{(\Xc)}$ that for some specified accuracy value $\epsilon \geq 0$ achieves 
\begin{equation} \label{eq:objective}
	\min_{\secvar \in \secset} \E_{\vx \sim \Pc^{(T)}}[f(\vx, \secvar)] \geq \tau^* - \epsilon.
		\vspace{-0.2em}
\end{equation} 
Besides achieving \eqref{eq:objective}, our goal is also to minimize the total number of required samples $T$.


We note that our optimization goal is different from the one of computing {\em deterministic} (pure strategy) solution $\vx \in \Xc$ and competing against $\tau = \max_{\vx \in \Xc} \min_{\secvar \in \secset} f(\vx, \secvar)$ as considered in \citep{bogunovic2018adversarially}. Our goal is to discover a {\em randomized} strategy and compete against $\tau^* \geq \tau$, which can be arbitrarily larger than $\tau$. Hence, mixed strategies considered in this work can provide arbitrarily better expected performance than such deterministic ones. Conceptually, randomization allows the decisions to be less predictable, and is a key feature necessary in many applications including security games \citep{sinha2018security}, adversarial learning \citep{vorobeychik2014} and sensing~\citep{krause2011robust}. This is also the case in the autonomous driving scenario considered in Section~\ref{sec:autonomous_driving}, where we show that deterministic strategies can be overly conservative. \looseness=-1
Finally, we also note that the same objective~\eqref{eq:maximin} is considered in~\citep{chen2017robust}, in the case of \emph{known} reward functions $f_i(\cdot) := f(\cdot, \vtheta_i)$, and $i \in \lbrace 1, \dots, m \rbrace$.

\textbf{Our Model.} 
We assume that the unknown objective $f$ is fixed and belongs to a Reproducing Kernel Hilbert Space (RKHS) $\mathcal{H}_k(D)$ corresponding to a positive semi-definite kernel function $k(\cdot, \cdot):D \times D \rightarrow \R$. Furthermore, we require $f$ to have a bounded RKHS norm, i.e., $\|f \|_k =\sqrt{\langle f, f \rangle_k\textbf{}} \leq B$ where $\| \cdot \|_k$ stands for the RKHS norm and $B$ is a known positive constant. The RKHS norm represents a measure of smoothness of $f$ as measured by the corresponding kernel. We note that these are the standard assumptions used in GP optimization (see, e.g.,~\citep{srinivas2009gaussian, chowdhury17kernelized, bogunovic2018adversarially}). 

For the kernel function, we assume $k\big((\vx, \secvar),(\vx,\secvar)\big)\leq 1$ for all $(\vx, \secvar) \in D$, which is without loss of generality if appropriate re-scaling is applied. Our setup also allows for composite kernels that can be constructed by using individual kernels $k_1: \Xc \times \Xc \rightarrow \R$ and $k_2: \secset \times \secset \rightarrow \R$, to obtain, for example, \emph{additive kernel} $k\big((\vx,\secvar),(\vx',\secvar') \big):= k_1(\vx,\vx') + k_2(\secvar, \secvar') $ or \emph{product kernel} $k\big( (\vx,\secvar),(\vx',\secvar') \big):= k_1(\vx,\vx') \cdot k_2(\secvar,\secvar')$.~Popularly used kernels are linear, squared exponential (SE) and Matérn:\looseness=-2 
\begin{align*}
k_{\text{Lin}}(\vx,\vx') &= \vx^T \vx', \\
k_{\text{SE}}(\vx,\vx') &= \exp \left(- \tfrac{1}{2l^2}\|\vx - \vx'\|^2 \right), \text{and} \\
k_{\text{Mat}}(\vx,\vx') &= \tfrac{2^{1-\nu}}{ \Gamma(\nu) } \bigg( \tfrac{\sqrt{2\nu} \|\vx - \vx'\| }{l} \bigg) J_{\nu}\bigg( \tfrac{\sqrt{2\nu} \|\vx - \vx'\| }{l} \bigg), 
\end{align*}
where $l$ is the length-scale parameter and $\nu > 0$ is a parameter that determines the smoothness \citep{rasmussen2006gaussian}.

Under such assumptions, the uncertainty over $f$ is naturally modeled as a Gaussian process $\mathrm{GP}(0, k((\vx, \secvar),(\vx',\secvar')))$. Further on, a Gaussian likelihood model for the observations can be used assuming the noise $\xi_t = y_t - f(\vx_t, \secvar_t)$ is drawn, independently across $t$, from $\mathcal{N}(0, \lambda)$. Here, $\lambda$ denotes a free hyper-parameter that may differ from the true noise variance $\sigma^2$. With this model in place, conditioned on the history of inputs  $\lbrace (\vx_1, \secvar_1), \dots, (\vx_t, \secvar_t) \rbrace$ and their noisy observations $\lbrace y_1, \dots, y_t \rbrace$, the posterior distribution under this prior is also Gaussian with the closed form posterior mean and variance:
\begin{align}
	\mu_{t}(\vx,\secvar) &= \vk_t(\vx,\secvar)^T\big(\vK_t + \lambda \mathbf{I}_t \big)^{-1} \vy_t,  \label{eq:posterior_mean} \\ 
	\sigma_{t}^2(\vx,\secvar) &= k((\vx, \secvar),(\vx,\secvar))  \nonumber \\
	&\quad-\vk_t(\vx,\secvar)^T \big(\vK_t + \lambda \mathbf{I}_t \big)^{-1} \vk_t(\vx,\secvar), \label{eq:posterior_variance}
\end{align}
s.t. $\vk_t(\vx,\secvar) = \big[k\big((\vx_j,\secvar_j),(\vx,\secvar)\big)\big]_{j=1}^t$, and $\vK_t = \big[k \big((\vx_j,\secvar_j),(\vx_{j'},\secvar_{j'})\big)\big]_{j,j'}$ is the kernel matrix. As described bellow, we make use of this model in our algorithm to sequentially learn about the unknown objective function.

\vspace{-0.7em}
\section{Proposed Algorithm and Theory}
\vspace{-0.7em}
Our algorithm, \ouralg, is shown in Algorithm~\ref{alg:iro}. It can be interpreted as a zero-sum game between a simulated adversary and a learner. The adversary plays actions from the set $\secset$, while the learner plays actions from $\Xc$. Because the true reward function $f(\cdot,\cdot)$ is unknown, the algorithm maintains and makes use of the optimistic upper confidence bound $\oucb_{t}(\cdot, \cdot)$ (defined below) of the unknown reward function. We define the confidence bounds as follows:
\begingroup\setlength{\abovedisplayskip}{3pt}
\setlength{\belowdisplayskip}{3pt}
\begin{align} 
    \ucb_{t}(\vx,\secvar) &:= \mu_t(\vx,\secvar) + \beta_{t+1} \sigma_{t}(\vx,\secvar) \label{eq:ucb}\\
    \lcb_{t}(\vx,\secvar) &:= \mu_t(\vx,\secvar) - \beta_{t+1} \sigma_{t}(\vx,\secvar), \label{eq:lcb}
\end{align}
where $\beta_t$ is the \emph{confidence parameter} that we set according to Lemma~\ref{conf_lemma} bellow. We also define their truncated versions: 
\begin{align}
    \oucb_{t}(\vx,\secvar) &:= \min \lbrace \ucb_{t}(\vx,\secvar), 1 \rbrace \label{eq:oucb}\\
    \olcb_{t}(\vx,\secvar) &:= \max \lbrace \lcb_{t}(\vx,\secvar), 0 \rbrace, \label{eq:olcb}
\end{align}
\endgroup
which we use in our algorithm. At every round $t$, \ouralg simulates the adversary by selecting a distribution over the $m$ values of $\vtheta$, i.e., $\vw_t \in \lbrace \vw \in [0,1]^{m}: \sum_{i=1}^{m} \vw[i] = 1 \rbrace$, where $\vw_t[i]$ denotes the probability of selecting $\secvar_i$. Subsequently, the learner \emph{best responds} by selecting $\vx_t$ based on the knowledge of $\vw_t$. After $T$ iterations, \ouralg returns the uniform distribution over $\lbrace \vx_1, \ldots, \vx_T\rbrace$, denoted with $\mathcal{U}^{(T)}$. Next, we explain how $\vw_t$ and $\vx_t$ are chosen in Algorithm~\ref{alg:iro}.

The multiplicative weight updates (MWU) rule \citep{freund1997} is used to select $\vw_t$ at every round $t$. 
We note that this algorithm is an online learning \emph{no-regret} algorithm that requires \emph{full-information} feedback at every round, i.e., observations that correspond to every pair $\lbrace (\vx_t, \secvar_i) \rbrace_{i=1}^{m}$. This is not possible in our setting where the learner only receives a single noisy observation that corresponds to the chosen pair $(\vx_t, \secvar_t)$. To cope with this, we make use of the upper confidence bound functions to effectively emulate the full information feedback.\footnote{A similar idea has recently been used by \citet{sessa2019noregret} in the context of multi-agent repeated games.} Hence, the MWU rule used in our algorithm is given by:
\vspace{-0.3em}
\begin{equation} \nonumber
    \vw_t[i] \propto \exp \Big\lbrace -\eta_T \sum_{j=1}^{t-1} \oucb_{j-1}(\vx_j, \secvar_i)  \Big\rbrace,
    \vspace{-0.3em}
\end{equation}
where $\eta_T$ is the learning rate parameter that we set in Theorem~\ref{thm:main} bellow. Another equivalent way of writing this rule is via the following recursive update:
\begin{equation} \nonumber
    \vw_t[i] =  \frac{\vw_{t-1}[i] \cdot \exp\big(- \eta_T \cdot \oucb_{t-1}(\vx_{t-1}, \secvar_i)\big)}{\sum_{j=1}^{m} \vw_{t-1}[j] \cdot \exp(- \eta_T \cdot \oucb_{t-1}(\vx_{t-1}, \secvar_j) )} \text{ .}
        \vspace{-0.3em}
\end{equation}

The learner then observes $\vw_t$, and plays the best response 
$\vx_t$ that is obtained by using the upper confidence bound instead of the true unknown function:\vspace{-0.0em}
\begin{equation} \label{eq:x_t}
    \vx_t = \argmax_{\vx \in \Xc} \bigg(\sum_{i=1}^{m} \vw_t[i] \cdot \oucb_{t-1}(\vx,\secvar_i)\bigg).
    \vspace{-0.3em}
\end{equation}
Finally, the unknown function is queried at $(\vx_t, \secvar_t)$, where $\secvar_t$ is selected as the parameter value that has the highest uncertainty for the selected $\vx_t$, i.e.,\looseness=-1
        \vspace{-0.0em}
\begin{equation} \label{eq:i_t}
    \secvar_t \in \argmax_{\secvar \in \secset} \sigma_{t-1}(\vx_t, \secvar).
        \vspace{-0.2em}
\end{equation}
The observed data $(\vx_t, \secvar_t, y_t)$ is then used to update the model via~\eqref{eq:posterior_mean} and~\eqref{eq:posterior_variance}.

\vspace{-0.5em}
\subsection{Main result}
\vspace{-0.5em}
To characterize our regret bounds, we make use of a suitable measure of complexity of the function class, the so-called \emph{maximum information gain}. It has been introduced by~\citet{srinivas2009gaussian}, and subsequently used in many different works on Bayesian (GP) optimization. At time $t$, it is defined as
\begin{equation}
\label{eq:max_info_gain}
  \gamma_t = \max_{\lbrace (\vx_1,\secvar_1), \dots, (\vx_t,\secvar_t)\rbrace} \frac{1}{2} \log  \det(\mathbf{I}_t + \lambda^{-1}\vK_t),
\end{equation}
and is used to measure the reduction in uncertainty about $f$ after receiving $t$ noisy observations that correspond to $\lbrace (\vx_1,\secvar_1), \dots, (\vx_t,\secvar_t) \rbrace$. In the case $D \subset \R^d$, this kernel-dependent quantity is sublinear in $t$ for various kernel functions, e.g., $\mathcal{O}((\log t)^{d+1})$ for squared exponential and $\mathcal{O}(t^{(d+1)d /((d+1)d + 2\nu)}\log t)$ for the Mat\'ern kernel with $\nu > 1$ \citep{srinivas2009gaussian}.

We use the following well-known result in GP optimization~\citep{srinivas2009gaussian, chowdhury17kernelized}, that allows for construction of statistical confidence bounds around the unknown function. \looseness=-1
{\setlength{\abovedisplayskip}{3.5pt}
\setlength{\belowdisplayskip}{3.5pt}
\begin{lemma}
  \label{conf_lemma}
    Let $f \in \Hc_k(D)$ with $\| f \|_k \leq B$, and consider the sampling model 
    \[\hspace{1em}y_t = f(\vx_t,\secvar_t) + \xi_t \text{, where } \xi_t  
    \sim \mathcal{N}(0,\sigma^2).\] 
    If the confidence parameter is set to
    \begin{equation} \label{eq:standard_beta}
      \beta_t = B + \sigma \lambda^{-1/2} \sqrt{2(\gamma_{t-1} + \ln(1 / \delta))}, 
    \end{equation}
    the following holds for every $(\vx,\secvar) \in D$ and $t \geq 1$,
    with probability at least $1 - \delta$:
    \begin{equation}  
        |\mu_{t-1}(\vx,\secvar) - f(\vx,\secvar)| \leq  \beta_t \sigma_{t-1}(\vx,\secvar),  \label{eq:conf_bounds_std}
    \end{equation}
    where $\mu_{t-1}(\cdot,\cdot)$ and $\sigma_{t-1}(\cdot,\cdot)$ are given in \eqref{eq:posterior_mean} and \eqref{eq:posterior_variance} with $\lambda > 0$.
\end{lemma}
}
\begin{algorithm}[!t]
    \caption{\ouralg} \label{alg:iro}
    \begin{algorithmic}[1]
        \Require Sets $\secset$, $\Xc$, kernel $k$, parameters $\eta_T$, $\lbrace \beta_t \rbrace_{t \ge 1}$
        \For {$t = 1,2,\dotsc, T$}
        \State For every $i \in \lbrace 1, \ldots, m \rbrace$ set 
        \vspace{-0.6em}
            \begin{equation*}
            \vw_t[i] \propto \exp \Big\lbrace -\eta_T \sum_{j=1}^{t-1} \oucb_{j-1}(\vx_j, \secvar_i)  \Big\rbrace
            \end{equation*}
        \vspace{-1em}
        \State Set 
        \vspace{-0.8em}
            \begin{equation*}
            \vx_t \leftarrow \argmax_{\vx \in \Xc} \sum_{i=1}^{m} \vw_t[i] \cdot \oucb_{t-1}(\vx,\secvar_i)
             \end{equation*}
        \vspace{-1em}
        \State $\secvar_t \leftarrow \argmax_{\secvar \in \secset} \sigma_{t-1}(\vx_t, \secvar_i)$
        \State Observe $y_t = f(\vx_t, \secvar_t) + \xi_t$  
        \State Update $\mu_t(\cdot,\cdot)$ and $\sigma_t(\cdot,\cdot)$ according to \par \eqref{eq:posterior_mean} and \eqref{eq:posterior_variance} by including $\lbrace (\vx_t, \secvar_t, y_t) \rbrace$ 
        \EndFor
        \State \textbf{end for}
\Ensure Uniform distr. $\mathcal{U}^{(T)}$ over $\lbrace \vx_1, \dots, \vx_T \rbrace$.
\vspace{0.1em}
    \end{algorithmic}
\end{algorithm}

Given the definitions \eqref{eq:oucb}-\eqref{eq:olcb}, and by conditioning on the event~\eqref{eq:conf_bounds_std} in Lemma~\ref{conf_lemma} holding true we have:
\begin{equation} \label{eq:ordering}
    1 \geq  \oucb_{t}(\vx,\secvar) \geq f(\vx, \secvar) \geq \olcb_{t}(\vx,\secvar) \geq 0,
\end{equation}
for every pair $(\vx, \secvar) \in D$ and $t\geq1$.

Next, we state our main theorem in which we bound the performance of \ouralg.
All the proofs from this section are provided in the supplementary material.

\begin{theorem}
    \label{thm:main}
    Fix $B>0$, $\epsilon > 0$, $\delta \in (0,1)$, $m \in \mathbb{Z}^{+}$, $\lambda \geq 1$, and suppose the following holds
    \begin{equation} \nonumber
        T \geq \frac{1}{\epsilon^2} \left( \frac{\log(m)}{2} + \beta_T \sqrt{32\lambda \gamma_T \log (m)}  + 16 \beta_T^{2} \lambda \gamma_T \right),
    \end{equation}
    for some $T \in \mathbb{Z}^{+}$. For any $f:D \rightarrow [0,1]$, 
    such that
    $f \in \Hc_k(D)$ and $\|f \|_k \leq B$, \ouralg with $\beta_t$ set as in Lemma~\ref{conf_lemma} and $\eta_T = \sqrt{\frac{8 \log m}{T}}$ achieves 
    \begin{equation} \nonumber
        \min_{\secvar \in \secset} \E_{\vx \sim \mathcal{U}^{(T)}} [f(\vx,  \secvar)] \geq \max_{\Pc \in \Delta{(\Xc)}} \min_{\secvar \in \secset} \E_{\vx \sim \Pc}[f(\vx, \secvar)] - \epsilon,
    \end{equation}
    after $T$ rounds with probability at least $1 - \delta$, where $\mathcal{U}^{(T)}$ is the distribution returned by \ouralg.
\end{theorem}

Our analysis is based on the regret bounding techniques for zero-sum games similarly to \citep{chen2017robust} (we bound the rate of convergence to an equilibrium of the game simulated by \ouralg), but with additional non-trivial challenges to characterize the excess regret due to the fact that $f$ is unknown. The result in this theorem holds for general kernels and it can be made more specific by substituting the bounds on $\gamma_T$ for different kernels. For example, for $D \subset \R^d$ and the widely used squared exponential kernel, we obtain $T = \mathcal{O}^* \big(\tfrac{1}{\epsilon^2}\big(\log(m) + (\log \tfrac{1}{\epsilon})^d\sqrt{\log(m)} + (\log \tfrac{1}{\epsilon})^{2d}\big)\big)$, for constant $\lambda, \sigma, B, d, m$, where $\mathcal{O}^*(\cdot)$ is used to hide dimension-independent $\log$ factors. In the same setting, \stableopt\citep{bogunovic2018adversarially} requires $T = \mathcal{O}^* \big(\tfrac{1}{\epsilon^2}\big(\log \tfrac{1}{\epsilon})^{2d}\big)$ samples to discover a \emph{deterministic} maximin strategy that is near-optimal with respect to a generally \emph{weaker} benchmark. 
Finally, in comparison to the result of~\citet{chen2017robust} where $T=\mathcal{O} \big( \tfrac{\log(m)}{\epsilon^2}\big)$ and $f$ is assumed to be \emph{known}, our bound characterizes an additional number of samples required for estimating the unknown RKHS function.

\vspace{-0.2em}
\subsubsection{Trading Off Worst-Case and Average-Case Performance}
\label{sec:trading_off}
\vspace{-0.5em}

In many scenarios, one might care about the performance of the reported distribution in the worst-case while also ensuring a good performance on ``average". A natural problem to consider is to trade off these two quantities by using the following objective:
\begin{equation*}
    W(\Pc):=(1-\chi) \cdot \mathop{\E}_{\substack{\secvar \sim \Qc \\ \vx \sim \Pc}}[f(\vx,\secvar)] + \chi \cdot \min_{\secvar \in \secset} \mathop{\E}_{\vx \sim \Pc}[f(\vx,\secvar)] \, ,
\end{equation*}
for some fixed distribution $\Qc \in \Delta(\secset)$ (e.g., the uniform distribution) and trade-off parameter $\chi \in (0,1]$. Note that by setting $\chi=1$, we recover the worst-case objective. Hence, our goal is to output $\Pc^{(T)} \in \Delta(\Xc)$ after $T$ rounds, such that for some  accuracy $\epsilon > 0$ 
\begin{equation} \label{eq:wc_avg_tradeoff_goal}
    W(\Pc^{(T)}) \geq W(\Pc^*) - \epsilon,
\end{equation}  
where 
$\Pc^* \in \argmax_{\Pc \in \Delta (\mathcal{X})} W(\Pc)$.

Extending our algorithm to this case amounts to modifying the best response rule (Line 3 of Algorithm~\ref{alg:iro}) as:\looseness=-1
\begin{align}\label{eq:modified_best_response}
    \vx_t = \argmax_{\vx \in \Xc} \Big[ & (1- \chi)\cdot \mathop{\E}_{\secvar \sim \Qc} [\oucb_{t-1}(\vx,\secvar)]    \nonumber \\[-2pt]
   & + \chi \cdot \sum_{i=1}^m \vw_t[i] \cdot \oucb_{t-1}(\vx,\secvar_i)\Big].
   \vspace{-0.5em}
\end{align}
The theoretical guarantees of \ouralg in this setting with the best-response rule as given in~\eqref{eq:modified_best_response} are provided in the following corollary.
{\setlength{\abovedisplayskip}{2pt}
\setlength{\belowdisplayskip}{2pt}
\begin{corollary}\label{corollary}
Let $Q$ be a fixed distribution in $\Delta(\Theta)$ and let $\chi \in (0,1]$ be a trade-off parameter. 
Under the setup of Theorem~\ref{thm:main}, and when the following holds
    \begin{equation} \nonumber
        T \geq \frac{1}{\epsilon^2} \left( \tfrac{\chi^2 \log(m)}{2} + \chi  \beta_T \sqrt{32\lambda \gamma_T \log (m)}  + 16 \beta_T^{2} \lambda \gamma_T \right),
    \end{equation}
    for some $T \in \mathbb{Z}^{+}$, \ouralg with best-response rule as in~\eqref{eq:modified_best_response}, achieves  
    \begin{equation} \nonumber
        W(\mathcal{U}^{(T)}) \geq W(\Pc^*) - \epsilon,
    \end{equation}
    after $T$ rounds with probability at least $1 - \delta$, where $\mathcal{U}^{(T)}$ is the returned uniform distribution over the queried points $\lbrace \vx_1, \dots, \vx_T \rbrace$.
\end{corollary}
}
The proof closely follows the one of Theorem~\ref{thm:main}. When $\chi = 1$, we recover Theorem~\ref{thm:main}, while the performance clearly improves for smaller values of $\chi$, i.e., when $\chi \in (0,1)$. We also note that for $\chi = 0$, our algorithm solves the stochastic optimization problem, and achieves the standard regret bound (as in~\citep{srinivas2009gaussian}) which is known to be nearly optimal for various kernels (see~\citep{scarlett2017lower}).

\vspace{-0.5em}
\section{Experiments}
\vspace{-0.5em}
In this section, we evaluate the performance of \ouralg on synthetic benchmarks and demonstrate the applicability of \ouralg in planning safe trajectories for autonomous vehicles guided by user's preferences. 

\vspace{-0.5em}
\subsection{Synthetic Experiments}
\vspace{-0.5em}
For a function $f: \mathcal{X} \times \secset \rightarrow \mathbb{R}$, we compute the  \emph{performance} of a mixed strategy $\mathcal{P}^{(T)} \in \Delta(\mathcal{X})$ as:
\begin{equation}
\label{eq:experiments_performance_indicator}
\min_{\secvar \in \secset} \mathbb{E}_{\vx \sim \mathcal{P}^{(T)}} \big[ f(\vx,\secvar) \big] \,.
\end{equation}
In case the strategy is deterministic $\vx_T \in \Xc$, the performance is computed by considering the Dirac distribution centered at $\vx_T$.
We compare the performance of 
\ouralg with the following baselines:

\vspace{-1em}
\begin{itemize}[leftmargin=4mm]
\setlength\itemsep{-0.1em}
\item   \textsc{StableOpt} \citep{bogunovic2018adversarially} searches for the deterministic max-min point.
\item   \textsc{GP-UCB} \citep{srinivas2009gaussian} seeks for a non-robust global optimum and selects $(\vx_t,\secvar_t) = \arg\max_{(\vx,\secvar) \in \mathcal{X}\times \secset} \oucb_{t-1}(\vx,\secvar)$ at every $t$. After $T$ iterations, we consider $\vx_T$ to be the returned point.\looseness=-1
\item  \textsc{RandMaxMin} selects the point reported by \textsc{StableOpt} \emph{or} \textsc{GP-UCB} with equal probability at every round, and returns a uniform distribution over these points.
\end{itemize}
\vspace{-1em}

\begin{figure*}[ht]
\begin{subfigure}[b]{0.47\textwidth}
\includegraphics[width=1.05\textwidth]{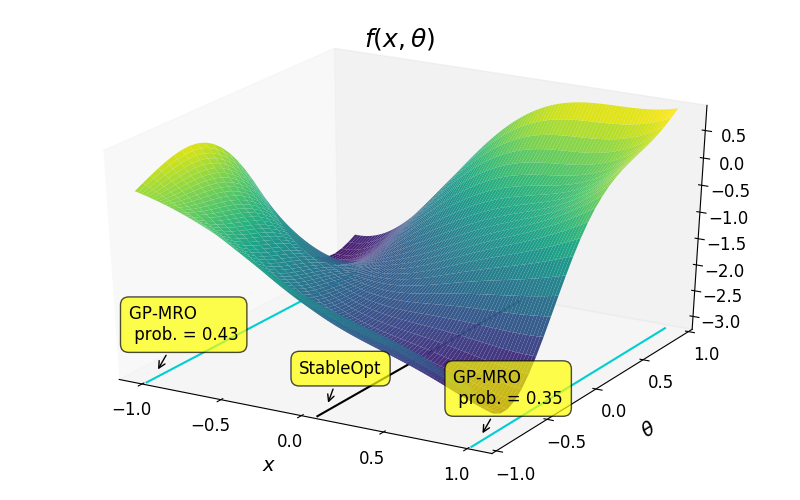}
\vspace{-1.8em}
\caption{Selected points}
\label{fig:random_RKSH:fig1}
\end{subfigure}
\hspace{1.8em}
\begin{subfigure}[b]{0.47\textwidth}
\includegraphics[width=\textwidth]{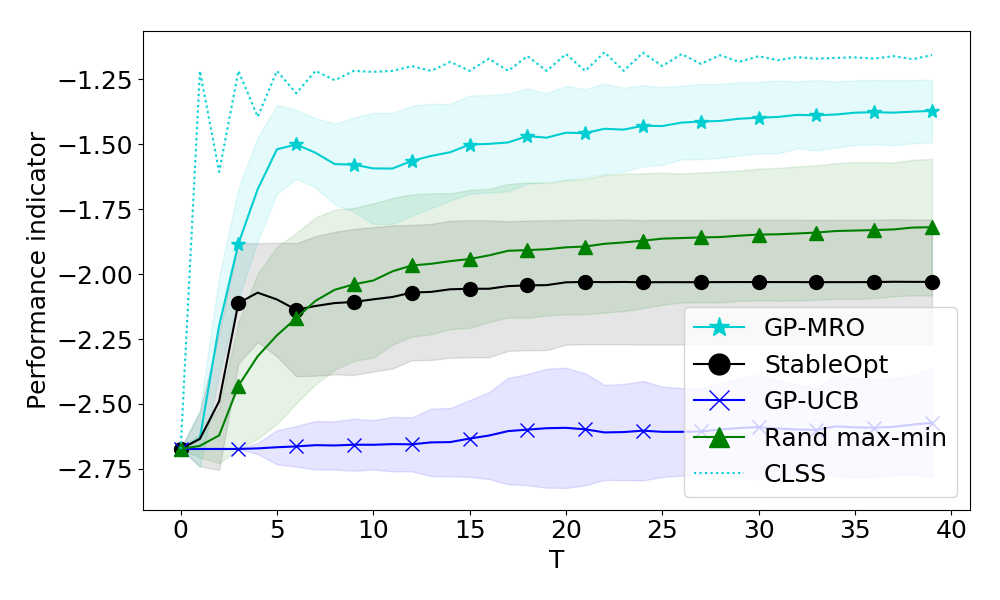}
\vspace{-2em}
\caption{Performance comparison}
\label{fig:random_RKSH:fig2}
\end{subfigure}
\vspace{-0.8em}
\caption{\small (a) Strategies returned by \textsc{StableOpt} and \ouralg after $T = 40$ iterations. (b) Comparison of the performance of the considered baselines, computed as in \eqref{eq:experiments_performance_indicator}. The proposed \ouralg algorithm outperforms all the baselines. \textsc{CLSS}\citep[Algorithm 1]{chen2017robust} assumes oracle access to $f$ and upper bounds the achievable results.}
 \vspace{-1em}
\end{figure*}

We set $\beta_T = 2.0$ for each of the above algorithms (we found the theoretical choice to be overly conservative, as also noted in previous works \citep{srinivas2009gaussian,bogunovic2018adversarially}), while $\eta_T$ is set according to Theorem~\ref{thm:main}.
As an idealized benchmark, we also test against \citep[Algorithm 1]{chen2017robust} (which we name via the authors’ surnames as CLSS) which assumes \emph{oracle} access to $f$ and thus upper bounds the achievable performance.

In the first experiment, we let $\mathcal{X}, \secset \subset [-1,1]$ with $|\mathcal{X}| = 100$, and $|\secset| = 30$, and sample a random function $f:\mathcal{X}\times  \secset \rightarrow \mathbb{R}$ from a $\mathrm{GP}(0,k)$ with kernel $k = k_{\text{Lin}} \cdot k_{\text{SE}}$. Moreover, we run the different baselines with the true prior $\mathrm{GP}(0,k)$ and noise standard deviation $\sigma=1.0$.

In Figure~\ref{fig:random_RKSH:fig1}, we show $f$ as well as the strategies returned by \textsc{StableOpt} and \ouralg after $T=40$ iterations. 
\textsc{StableOpt} converges to the max-min point of $f$,
while the distribution returned by \ouralg assigns most of the probability mass to points $x= +1.0$ and $x = -1.0$. As shown in Figure~\ref{fig:random_RKSH:fig2}, this leads to a higher performance compared to all the considered baselines. \looseness=-1

Next, we consider the synthetic function $g_\text{poly} : \R^2 \rightarrow \R$ from \citep{bertsimas2010robust},
and the robust optimization task from~\citep{bogunovic2018adversarially}. The goal is to select points $\vx = (x_1,x_2)$ 
that maximize $g_\text{poly}$ subject to the worst-case perturbation $\secvar \in \secset$. We map such problem to our setting by defining $f(\vx,\secvar) = g_\text{poly}(\vx-\secvar)$.
The decision space $\Xc$ consists of a uniformly spaced grid of $10'000$ points, while the set of perturbations $\secset$ is obtained by drawing $100$ random points from the unit ball centered at the origin. 

\begin{figure*}[t]
\begin{subfigure}[b]{0.47\textwidth}
\includegraphics[width=1.05\textwidth]{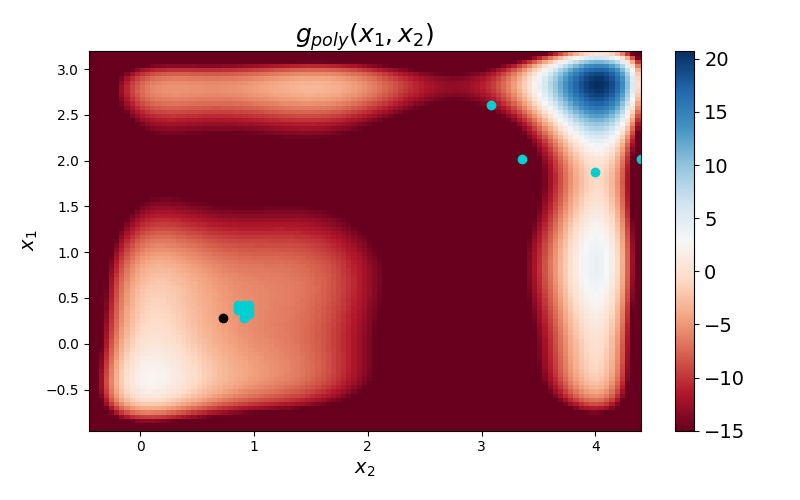}
\vspace{-2.0em}
\caption{Selected points}
\label{fig:f_bertsimas:fig1}
\end{subfigure}
\hspace{1.8em}
\begin{subfigure}[b]{0.47\textwidth}
\includegraphics[width=\textwidth]{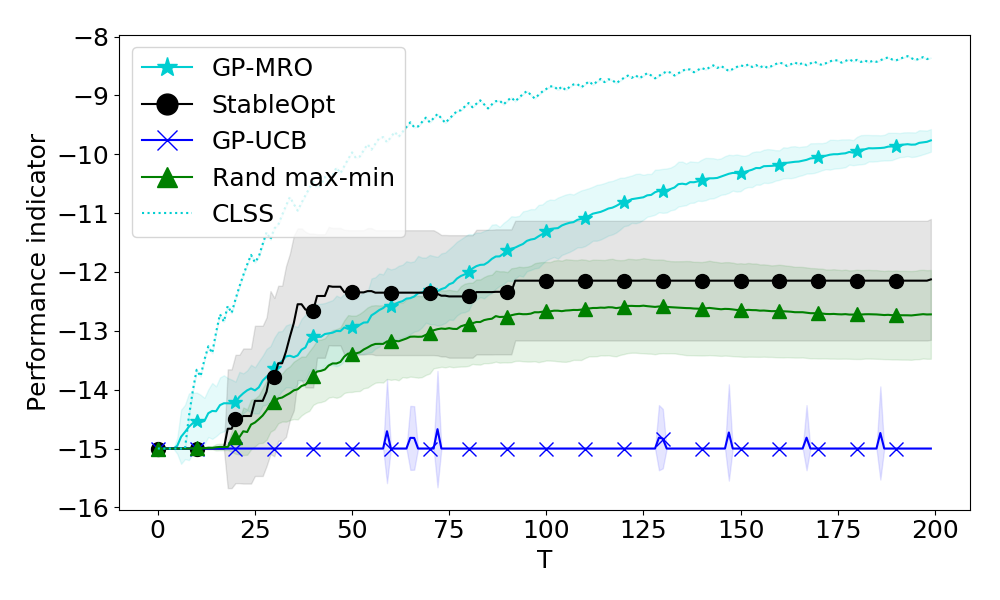}
\vspace{-2.0em}
\caption{Performance comparison}
\label{fig:f_bertsimas:fig2}
\end{subfigure}
\vspace{-0.8em}
\caption{\small (a) Supports of the strategies returned by \textsc{StableOpt} (in black) and \ouralg (in cyan) after $T = 200$ iterations. \textsc{StableOpt} reports a deterministic strategy, while \ouralg returns a randomized strategy. (b) Performance of the different baselines, computed as in \eqref{eq:experiments_performance_indicator}. The mixed strategy returned by \ouralg outperforms all the baselines. The \textsc{CLSS} algorithm has oracle access to $f$ and upper bounds the achievable performance.}
  \vspace{-0.5em}
\end{figure*}

We set noise standard deviation $\sigma = 1.0$ and run all the algorithms using Matérn kernel $k_\text{Mat}$ for $T=200$ iterations 
(kernel hyperparameters are found via maximum-likelihood method). 
In Figure~\ref{fig:f_bertsimas:fig1}, we plot the function $g_\text{poly}$ as well as the support of the strategies returned by \textsc{StableOpt} (in black) and \ouralg (in cyan). For \ouralg we plot only points selected with probability mass greater than $0.01$. \textsc{StableOpt} is able to discover the max-min point of $g_\text{poly}$, while \ouralg randomizes between points in the max-min region and points close to the global optimum. This leads to a higher performance compared to other baselines, as shown in Figure~\ref{fig:f_bertsimas:fig2}.\looseness=-1

\vspace{-0.5em}
\subsection{Human-assisted trajectory planning for autonomous vehicles}
\label{sec:autonomous_driving}
\vspace{-0.5em}

\begin{figure}[t]
\hspace{-0.7em}
\begin{subfigure}[b]{0.50\textwidth}
\includegraphics[width=1.0\textwidth,  angle =0 ]{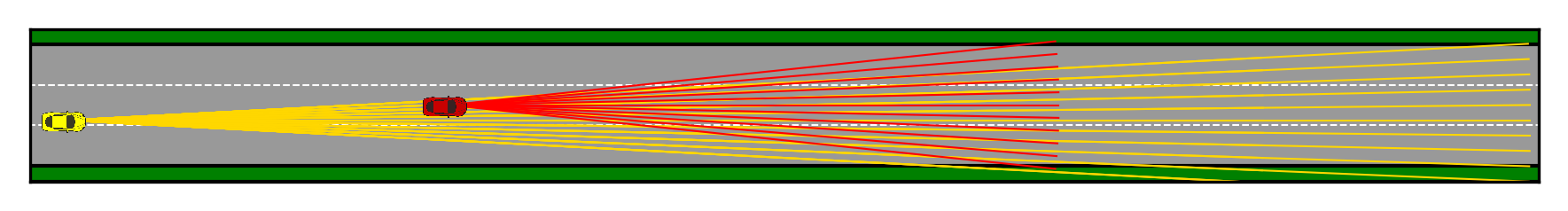}
\vspace{-1.7em}
\caption{Initial plausible trajectories}
\label{fig:driving:initial_trajs}
\includegraphics[width=1.0\textwidth]{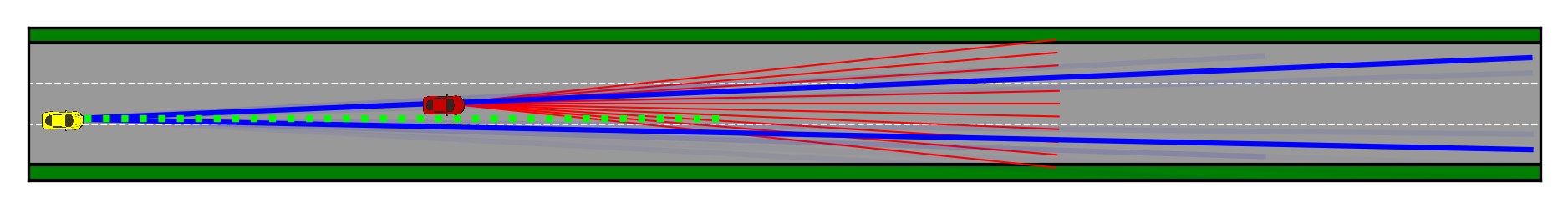}
\vspace{-1.7em}
\caption{Computed strategies, $\chi = 1$}
\label{fig:driving:optimal_trajs_lam_1}
\includegraphics[width=1.0\textwidth]{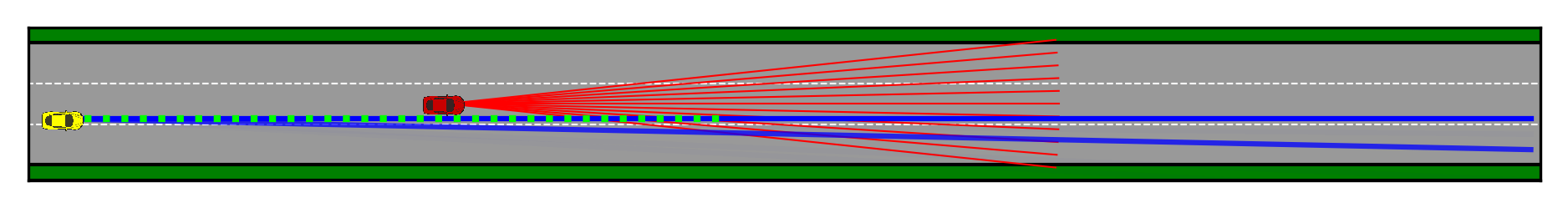}
\vspace{-1.7em}
\caption{Computed strategies, $\chi = 0.8$}
\label{fig:driving:optimal_trajs_lam_08}
\vspace{-0.5em}
\end{subfigure}
\caption{\small(a) Initial plausible trajectories of the AV (yellow) and the HV (red). In (b) and (c), the robust deterministic strategy (in dotted light-green) corresponds to breaking and not overtaking. The mixed strategy $\mathcal{U}^{(T)}$ found by \ouralg is represented by the blue trajectories (intensities proportional to their probabilities) and depends on the trade-off parameter $\chi$.}
\vspace{-1.0em}
\end{figure}

We study the problem of planning safe trajectories for an Autonomous Vehicle (AV) driving on roads shared with human-driven vehicles (HVs).
We consider the situation depicted in Figure~\ref{fig:driving:initial_trajs}, where  the AV (in yellow) is approaching, with a speed of $\SI[per-mode=symbol]{20}
{\meter\per\second}$, a HV (in red) driving at a constant speed of $\SI[per-mode=symbol]{10}
{\meter\per\second}$. The intentions of the HV are uncertain and this should be taken into account when planning the AV's trajectory. 

In the context of autonomous driving and AV-HV interactions, deterministic strategies would make AVs' actions predictable, hence giving a significant advantage to HVs. We observe this fact in our simulations, where such strategies tend to be overly conservative and prevent the AV from completing the overtake manoeuvre. Similarly, we expect this to occur in many other challenging scenarios such as intersections \citep{liu2018intersection}, or when merging into dense lanes \citep{bouton2019merging}.
Instead, we model such problem according to Section~\ref{sec:problem_formulation} and seek for robust \emph{mixed strategies} for the AV. This is in contrast with previous works (e.g., \citep{fisac2018hierarchical, sadigh2016planning}) where deterministic strategies are found, assuming a specific behavioral model for the HV.\looseness=-1 

Further on, our goal is to plan trajectories for the AV which best reflect typical \emph{human driving preferences} (e.g., driving styles, security measures, and safe behaviors that the AV should follow).
For instance, in the specific situation of Figure~\ref{fig:driving:initial_trajs}, a good trajectory for the AV should depend on the importance that humans give to overtaking rather than breaking behind the HV. 
We encode such driving preferences with an unknown \emph{scoring function}. We assume we can learn such function by sequential evaluations obtained interacting with a \emph{user} who assists our planning phase.

Computing such mixed strategies requires enough computation and relies on sequential interactions with the user. Hence, after illustrating our approach, we propose an \emph{offline} scheme to pre-compute a control \emph{policy} for the AV using \ouralg.

\textbf{Decision sets.} A strategy for the AV consists of selecting a steering angle $x_1\in [-\tfrac{\pi}{60},\tfrac{\pi}{60}]$, and an acceleration $x_2 \in [-10,1]$. Once chosen, both are assumed to be constant for the horizon of $\SI[per-mode=symbol]{8}
{\second}$. Hence, we let $\Xc$ be the set of points $\vx = (x_1, x_2)$. Similarly, we assume the HV travels at a constant speed and can choose a steering angle $\secvar \in \secset = [-\tfrac{\pi}{30},\tfrac{\pi}{30}]$. We discretize both $\mathcal{X}$ and $\secset$ using uniform grids of $121$ and $11$ points, respectively. Car trajectories (depicted in Figure~\ref{fig:driving:initial_trajs}) are computed using the commonly used discrete-time bicycle model \citep{polack2017bycicle} with time steps of $\SI[per-mode=symbol]{0.04}{\second}$.\looseness=-1

\textbf{Optimization goal.} We let the scoring function $f:\mathcal{X} \times \secset \rightarrow [0,1]$ reflect the humans' driving preferences for the AV. As discussed later, $f$ measures how rewarding is for the AV to select a possible $\vx \in \Xc$ when the HV decides to steer with angle $\secvar \in \secset$.
Our goal is to compute a robust mixed strategy which solves the problem in~\eqref{eq:maximin}.
More generally, according to Section~\ref{sec:trading_off}, we can incorporate \emph{priors} $\mathcal{Q} \in \Delta(\secset)$ on HV's behaviors and find strategies that can trade-off worst-case and average-case performance, for a trade-off parameter $\chi \in (0,1]$. 

\textbf{Scoring function.} We assume that $f$ is initially unknown but can be learned by iteratively querying the user. Querying $f$ at a given point $(\vx,\secvar)$ consists of: 1) Forward simulating the AV's and HV's trajectories corresponding to $\vx$ and $\secvar$ and 2) Presenting the outcome of such simulation to the user who assigns a score to the considered trajectories.
In this experiment, we assume such score is determined by a feature vector $\mathbf{z}=[z_1, z_2, z_3]\in \R^3$ that can be extracted from the simulated trajectories. Such vector consists of: longitudinal distance travelled by the AV ($z_1$), AV's maximum absolute lateral position ($z_2$), and the minimum distance between the AV and the human-driven car ($z_3$).
We use a model of the unknown $f$ of the following form: 
$f_\text{p}(z_1) + f_\text{r}(z_2) + f_\text{eb}(z_3)$, where $ f_\text{p}$ rewards progress, $f_\text{r}$ penalizes exiting the road limits, while $ f_\text{eb}$ penalizes the AV if it gets too close 
to the human-driven car and therefore needs to activate emergency breaking. In future work, we plan to replace our model and test our approach with scores coming from real users. \looseness=-1

\begin{SCfigure*}
\hspace{1em}
\begin{subfigure}[b]{0.29\textwidth}
\includegraphics[width=0.95\textwidth]{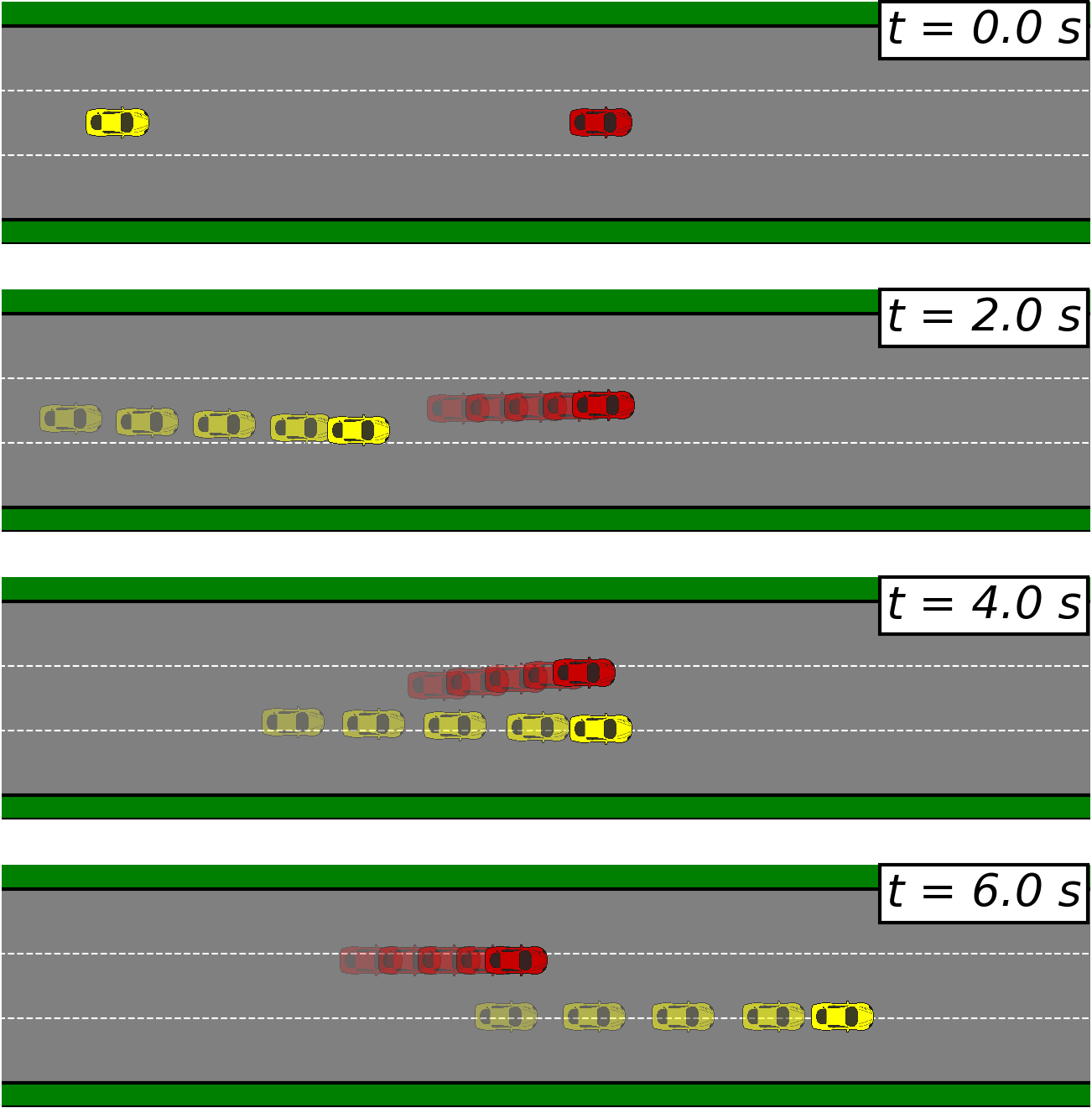}
\caption{\ouralg}
\end{subfigure}
\hspace{0.5em}
\begin{subfigure}[b]{0.29\textwidth}
\centering
\includegraphics[width=0.95\textwidth]{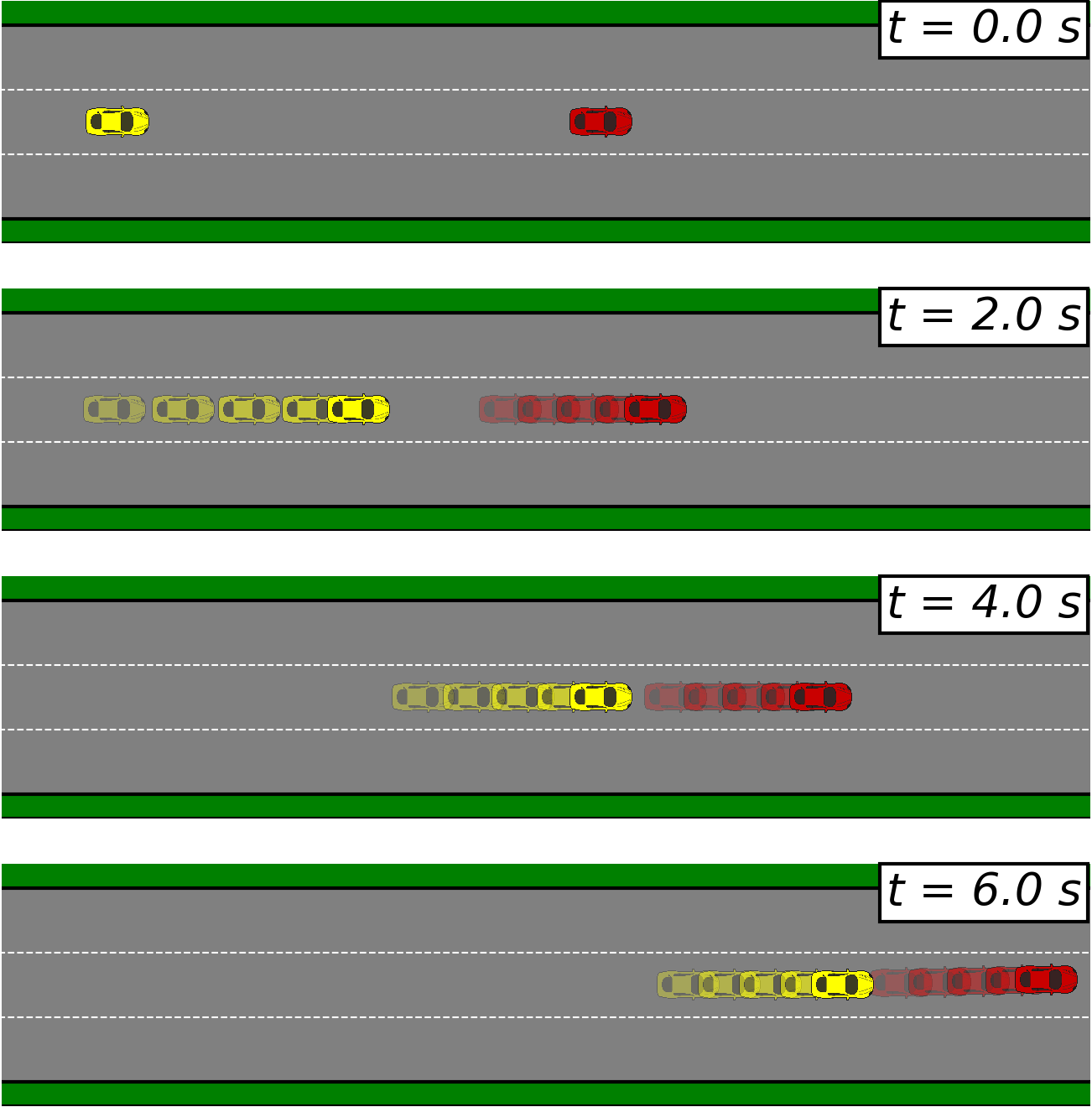}
\caption{Max-min deterministic strategy}
\end{subfigure}
\hspace{0em}
  \caption{ Closed-loop simulation of the AV (yellow) and human-driven car (red). At every iteration, the AV implements (a) the randomized policy found by  \ouralg or (b) the deterministic max-min strategy. The human-driven car follows the noisy rational Boltzmann policy \eqref{eq:rational_boltzmann}. The robust deterministic strategies are overly conservative, while \ouralg algorithm allows the AV to safely overtake. \vspace{0.0em}}
\label{fig:driving:closed_loop_snaps}
\vspace{-0.8em}
\end{SCfigure*}

\vspace{-0.5em}
\subsection{Illustration of the mixed strategies computed by \ouralg}
\vspace{-0.5em}
We consider the configuration in Figure~\ref{fig:driving:initial_trajs} and compute a mixed strategy for the AV running  \ouralg for $T = 100$ iterations. We set trade-off parameter $\chi = 1$, $\beta_T = 0.5$, and $\eta_T= 0.5$. To learn $f$, we fit a GP with kernel function $k(\mathbf{z},\mathbf{z}') = k_\text{Mat}^1(z_1, z_1') +  k_\text{Mat}^2(z_2, z_2') + k_\text{Mat}^3(z_3, z_3')$ where the feature vector $\mathbf{z}$ is computed as explained above. 
In Figure~\ref{fig:driving:optimal_trajs_lam_1}, we depict (in blue) the support of the mixed strategy $\mathcal{U}^{(T)}$ where the color intensity of a trajectory is proportional to its probability. Additionally, we show (in dotted light-green) the trajectory corresponding to the robust deterministic strategy $\vx_\text{r} \in \arg\max_{\vx \in \mathcal{X}} \min_{\secvar \in \secset} f(\vx,\secvar)$. The strategy $\mathcal{U}^{(T)}$ randomizes between an overtake from the left or the right side. 
Instead, $\vx_\text{r}$ amounts to breaking and thus never overtaking.

Our next goal is to find a strategy for the AV which can trade off the worst-case with average-case performance.
Let us assume that, with probability $0.2$, the HV doesn't realize the presence of the AV and thus has no intention to steer. In this case, we can seek for the optimal strategy for the AV by setting $\chi = 0.8$ and letting $\mathcal{Q}\in \Delta(\secset)$ be a Dirac distribution corresponding to the HV proceeding straight. In Figure~\ref{fig:driving:optimal_trajs_lam_08} we depict the strategy returned by \ouralg, together with the trajectory $\vx_\text{r} \in \arg \max_{\vx \in \mathcal{X}}  (1-\chi) \cdot \mathbb{E}_{\secvar \sim \mathcal{Q}} \big[ f(\vx,\secvar) \big] +  \chi \cdot \min_{\secvar \in \secset} f(\vx,\secvar) $. In this case, $\mathcal{U}^{(T)}$ favors an overtake from the right, while $\vx_\text{r}$ still leads to no overtaking. 

\vspace{-0.7em}
\subsection{Closed-loop simulations}
\vspace{-0.6em}
We propose the following \emph{offline} procedure to pre-compute a control \emph{policy} for the AV. We consider a finite set of $\sim 8'000$ possible scenarios $\mathbf{s} \in \mathcal{S} \subset \mathbb{R}^5$, each describing the initial and relative positions and velocities of the two cars. We compute a mixed strategy $\mathcal{U}^{(T)}(\mathbf{s})$ for each scenario $\mathbf{s} \in \mathcal{S}$ using \ouralg with $\chi = 1$.
Moreover, to make our approach more tractable, we query $f$ at chosen points $(\vx_t,\secvar_t)$ (Line 5 in Algorithm~\ref{alg:iro}) only if $\sigma_{t-1}(\vx_t,\secvar_t)$ is greater than $0.005$. By doing so, we end up with a policy mapping scenarios $\mathbf{s} \in \mathcal{S}$ to distributions $\mathcal{U}^{(T)}(\mathbf{s}) \in \Delta(\mathcal{X})$ after a total number of 136 queries of the unknown function. 

We evaluate the policy \emph{online}, in a receding-horizon fashion: Starting from given initial positions and velocities, every $\SI{2}{\second}$ we map the cars' positions and velocities to the closest $\mathbf{s} \in \mathcal{S}$ (using a nearest-neighbour tree-based algorithm) and let the AV sample its trajectory from $\mathcal{U}^{(T)}(\mathbf{s})$. For the behavior of the HV we implement a noisy rational Boltzmann policy (as in \citep{fisac2018hierarchical}) where, in a given scenario $\mathbf{s} \in \mathcal{S}$, $\secvar \in \secset$ is sampled with probability
\begin{equation}\label{eq:rational_boltzmann}
\mathbb{P}[\secvar = \secvar_i \mid \mathbf{s} \, ] \propto \exp \Big( { \mathbb{E}_{\vx \sim \mathcal{U}^{(T)}(\mathbf{s})} \:  f_{H}(\secvar_i, \vx)} \Big) \,.
\end{equation}
The function $f_{H}$ rewards progress for the HV and penalizes exiting the road or getting too close to the AV, the same way as $f$ does for the AV.

\begin{table}
{\small
\centering
\begin{tabular}{l | c | c}
                        & \small \ouralg & \small \makecell{ Deterministic \\ max-min }\\
                        \hline
\small \# of overtakes & \small $408/1000$  & \small $0/1000$      \\
\small avg. final pos. AV & \small  $\SI{169.4}{\meter}$  & \small    $\SI{123.1}{\meter}$  \\
\small avg. final pos. human   &  \small  $\SI{139.8}{\meter}$     &  \small   $\SI{139.9}{\meter}$    \\
\hline
\end{tabular}
\vspace{-0.3em}
\caption{Number of overtakes and cars' average final positions out of 1000 closed-loop simulations of $\SI{10}{\second}$.}
\label{table:num_overtakes}
\vspace{-1em}
}
\end{table}

In Figure~\ref{fig:driving:closed_loop_snaps}, we plot several snapshots of a closed-loop simulation of $\SI{10}{\second}$ where the AV samples trajectories from the pre-computed policy (a), and where the AV chooses the max-min strategy $\vx_\text{r}$ at every iteration (b). As can be seen from Figure~\ref{fig:driving:closed_loop_snaps}, the proposed approach allows the AV to safely overtake, while the robust deterministic strategy is too conservative and forces the AV to break behind the HV. We repeat the closed-loop simulation for $1'000$ times (for fixed initial positions and velocities of the two cars). As reported in Table~\ref{table:num_overtakes}, the deterministic strategy is non-overtaking and the AV reaches an average final longitudinal positions of $\SI{123.1}{\meter}$. 
Instead, using the pre-computed randomized policy the AV successfully overtakes the human-driven car in $408$ cases (in the remaining cases it breaks behind the HV), reaching an average final position of $\SI{169.4}{\meter}$.

\vspace{-0.5em}
\section{Conclusion}
\vspace{-0.5em}
We have studied a robust optimization problem in which the objective function is unknown and depends on an uncertain parameter. For this problem, we have proposed a novel sample-efficient algorithm \ouralg, which can discover a near-optimal randomized and robust strategy. We have established rigorous theoretical guarantees and designed a variant of \ouralg that effectively trades off worst-case and average-case performance. In synthetic experiments and trajectory planning tasks, we have showed that our proposed algorithm significantly outperforms existing baselines.

\subsubsection*{Acknowledgments}
This work was gratefully supported by the Swiss National Science Foundation, under the grant SNSF $200021$\textunderscore$172781$, by the European Union's ERC grant $815943$, and ETH Z\"urich Postdoctoral Fellowship 19-2 FEL-47.

\bibliography{mybib.bib}

\newpage
\onecolumn
\appendix

{\centering
    {\huge \bf Supplementary Material}
    
    {\Large \bf  Mixed Strategies for Robust Optimization of Unknown Objectives\\ [2mm] {\normalsize \bf {Pier Giuseppe Sessa, Ilija Bogunovic, Maryam Kamgarpour, Andreas Krause (AISTATS 2020)} \par }  
}}

\vspace*{-1ex}
\section{Proof of Theorem~\ref{thm:main}}
\vspace*{-1ex}

\begin{proof}
In this proof, we condition on the event in Lemma~\ref{conf_lemma} holding true, meaning that $\ucb_t$ and $\lcb_t$ provide valid confidence bounds as per~\eqref{eq:conf_bounds_std}.
As stated in the lemma, this holds with probability at least $1-\delta$.

Our main goal in this proof is to upper bound the difference: 
\begin{equation} \label{eq:regret_proof}
\max_{\Pc \in \Delta{(\Xc)}} \min_{\secvar \in \secset} \E_{\vx \sim \Pc}[f(\vx, \secvar)] - \min_{\secvar \in \secset} \frac{1}{T} \sum_{t=1}^{T} f(\vx_t, \secvar).
\end{equation}
To do so, we provide upper and lower bounds of the first and second terms, respectively, and then we upper bound their difference. 

First, we show that the following holds:
\begin{equation} \label{eq:second_term}
    \min_{\secvar \in \secset}  \frac{1}{T} \sum_{t=1}^{T} f(\vx_t, \secvar) \geq  
    \bigg(\min_{\secvar \in \secset} \frac{1}{T} \sum_{t=1}^{T} \oucb_{t-1}(\vx_t,\secvar)\bigg) - 4\beta_T \sqrt{\frac{\lambda \gamma_T}{T}},
\end{equation}
where $\vx_t$ is the point queried at time $t$.

To prove Eq.~\eqref{eq:second_term} we use the lower confidence bound and~\eqref{eq:ordering}:
\begin{align}
    \min_{\secvar \in \secset}  \frac{1}{T} \sum_{t=1}^{T} f(\vx_t, \secvar)  &\geq \min_{\secvar \in \secset}  \frac{1}{T} \sum_{t=1}^{T} \lcb_{t-1}(\vx_t, \secvar) \\
    &= \min_{\secvar \in \secset} \frac{1}{T} \sum_{t=1}^{T} \big( \ucb_{t-1}(\vx_t,\secvar) - 2\beta_t \sigma_{t-1}(\vx_t, \secvar)\big) \label{eq:second_term1} \\
    &\geq \bigg(\min_{\secvar \in \secset} \frac{1}{T} \sum_{t=1}^{T} \oucb_{t-1}(\vx_t,\secvar)\bigg) -  \max_{\secvar \in \secset} \frac{1}{T} \sum_{t=1}^{T} 2\beta_t \sigma_{t-1}(\vx_t, \secvar) \\
    &\geq \bigg(\min_{\secvar \in \secset} \frac{1}{T} \sum_{t=1}^{T} \oucb_{t-1}(\vx_t,\secvar)\bigg) - \frac{2\beta_T}{T} \sum_{t=1}^{T} \max_{\secvar \in \secset} \sigma_{t-1}(\vx_t, \secvar) \label{eq:second_term2} \\
    &= \bigg(\min_{\secvar \in \secset} \frac{1}{T} \sum_{t=1}^{T} \oucb_{t-1}(\vx_t,\secvar)\bigg) - \frac{2\beta_T}{T} \sum_{t=1}^{T} \sigma_{t-1}(\vx_t, \secvar_t) \label{eq:second_term3}\\
    &\geq \bigg(\min_{\secvar \in \secset} \frac{1}{T} \sum_{t=1}^{T} \oucb_{t-1}(\vx_t,\secvar)\bigg) - 4\beta_T \sqrt{\frac{\gamma_T \lambda}{T}},\label{eq:second_term4}
\end{align}
where \eqref{eq:second_term1} follows from the definition of the confidence bounds in \eqref{eq:ucb} and \eqref{eq:lcb}, \eqref{eq:second_term2} is due to monotonicty of $\beta_t$, and \eqref{eq:second_term3} is by rule~\eqref{eq:i_t} used in Algorithm~\ref{alg:iro} to select $\secvar_t$. Finally, \eqref{eq:second_term4} is obtained via the standard result from~\citep{srinivas2009gaussian, chowdhury17kernelized}
\begin{equation}
    \sum_{t=1}^T \sigma_{t-1}(\vx_t, \secvar_t) \leq \sqrt{4 T \lambda \gamma_T}, \label{eq:sigma_bound}
\end{equation}
when $\lambda \geq 1$.

Next, we show that the first term can be upper bounded as follows:
\begin{equation*}
    \max_{\Pc \in \Delta{(\Xc)}} \min_{\secvar \in \secset} \E_{\vx \sim \Pc}[f(\vx, \secvar)] \leq \frac{1}{T} \sum_{t=1}^T \E_{\secvar\sim \vw_t}[\oucb_{t-1}(\vx_t, \secvar)].
\end{equation*}
To prove this, we start by upper bounding the minimum value of the inner objective:
\begin{align}
    \max_{\Pc \in \Delta{(\Xc)}} \min_{\secvar \in \secset} \E_{\vx \sim \Pc}[f(\vx, \secvar)] &\leq \max_{\Pc \in \Delta{(\Xc)}} \frac{1}{T} \sum_{t=1}^T \sum_{i=1}^{m} \vw_t[i] \cdot \E_{\vx \sim \Pc}[f(\vx, \secvar_i)] \label{eq:first_term0}\\
    &\leq \max_{\Pc \in \Delta{(\Xc)}} \frac{1}{T} \sum_{t=1}^T \sum_{i=1}^{m} \vw_t[i] \cdot \E_{\vx \sim \Pc}\big[\oucb_{t-1}(\vx, \secvar_i)\big] \label{eq:first_term1} \\
    &= \max_{\Pc \in \Delta{(\Xc)}} \frac{1}{T} \sum_{t=1}^T\E_{\vx \sim \Pc}\bigg[ \sum_{i=1}^{m} \vw_t[i] \cdot \oucb_{t-1}(\vx, \secvar_i)\bigg] \label{eq:first_term2} \\
    &\leq \frac{1}{T} \sum_{t=1}^T \max_{\Pc \in \Delta{(\Xc)}} \E_{\vx \sim \Pc}\bigg[ \sum_{i=1}^{m} \vw_t[i] \cdot \oucb_{t-1}(\vx, \secvar_i)\bigg] \label{eq:first_term3} \\
    &= \frac{1}{T} \sum_{t=1}^T \max_{\vx \in \Xc} \sum_{i=1}^{m} \vw_t[i] \cdot \oucb_{t-1}(\vx, \secvar_i) \label{eq:first_term4} \\
    &= \frac{1}{T} \sum_{t=1}^T  \sum_{i=1}^{m} \vw_t[i] \cdot \oucb_{t-1}(\vx_t, \secvar_i)\,. \label{eq:first_term5}
    \end{align}

We obtain Eq.~\eqref{eq:first_term0} as the following trivially holds $$\min_{\secvar \in \secset} \E_{\vx \sim \Pc    }[f(\vx, \secvar)] \leq \sum_{i=1}^m \vw_t[i] \cdot \E_{\vx \sim \Pc}[f(\vx, \secvar_i)]$$ for each $t$ and $\vw_t \in \lbrace \vw \in [0,1]^{m}: \sum_{i=1}^{m} \vw[i] = 1 \rbrace$, and hence it also holds for the average value  $$\min_{\secvar \in \secset} \E_{\vx \sim \Pc    }[f(\vx, \secvar)] \leq \frac{1}{T} \sum_{t=1}^T  \sum_{i=1}^m \vw_t[i] \cdot \E_{\vx \sim \Pc}[f(\vx, \secvar_i)].$$
Eq.~\eqref{eq:first_term1} follows from~\eqref{eq:ordering},~\eqref{eq:first_term2} follows by the linearity of expectation, and~\eqref{eq:first_term4} 
holds since Dirac delta $\vdelta_{\vx}$, $\forall \vx \in \Xc$, is in $\Delta(\Xc)$. 
Finally, \eqref{eq:first_term5} follows by rule~\eqref{eq:x_t} used in Algorithm~\ref{alg:iro} to select $\vx_t$.

Next, we bound the difference in \eqref{eq:regret_proof} by combining the bounds obtained in \eqref{eq:second_term4} and \eqref{eq:first_term5}: \looseness=-1
\begin{align}
    \max_{\Pc \in \Delta{(\Xc)}} \min_{\secvar \in \secset}& \E_{\vx \sim \Pc}[f(\vx, \secvar)] - \min_{\secvar \in \secset} \frac{1}{T} \sum_{t=1}^{T} f(\vx_t, \secvar) \nonumber \\ 
    &\leq \frac{1}{T} \sum_{t=1}^T \E_{\secvar\sim \vw_t}\big[\oucb_{t-1}(\vx_t, \secvar)\big] - \bigg(\min_{\secvar \in \secset} \frac{1}{T} \sum_{t=1}^{T} \oucb_{t-1}(\vx_t,\secvar)\bigg) + 4\beta_T \sqrt{\frac{\gamma_T \lambda}{T}} \nonumber \\
    &\leq \sqrt{\frac{\log(m)}{2T}} + 4\beta_T \sqrt{\frac{\gamma_T \lambda}{T}} \label{eq:final_term},
\end{align}
where \eqref{eq:final_term} follows by the guarantees of the no-regret online multiplicative weight updates algorithm played by the adversary, that is,
\begin{equation}\label{eq:MWU_regret_bound}
    \frac{1}{T} \sum_{t=1}^T \E_{\secvar\sim \vw_t}\big[\oucb_{t-1}(\vx_t, \secvar)\big] - \bigg(\min_{\secvar\in \secset} \frac{1}{T} \sum_{t=1}^{T} \oucb_{t-1}(\vx_t,\secvar)\bigg) \leq \sqrt{\frac{\log(m)}{2T}},
\end{equation}
with the learning rate set to $\eta_T = \sqrt{\frac{8 \log(m)}{T}}$. For more details on this result see \citep[Section 4.2]{cesa-bianchi_prediction_2006} where the same online algorithm is considered. Specifically, the result above follows from~\citep[Theorem 2.2]{cesa-bianchi_prediction_2006} by noting that $\sum_{t=1}^T \E_{\secvar\sim \vw_t}\big[\oucb_{t-1}(\vx_t, \secvar)\big] = \sum_{t=1}^T \vw_t^T \cdot\oucb_{t-1}(\vx_t, \cdot)$, $\min_{\secvar \in \secset} \sum_{t=1}^{T} \oucb_{t-1}(\vx_t,\secvar)= \min_{\vw \in \Delta(\secset)} \sum_{t=1}^{T} \vw^T \cdot \oucb_{t-1}(\vx_t,\cdot)$ and $\oucb_{t-1}(\cdot,\cdot) \in [0,1]$ for every $t$. In our case, the objective function changes with $t$ but remains bounded, which allows the result to hold despite the changes (see time-varying games result extension~\citep[Remark 7.3]{cesa-bianchi_prediction_2006}). 

By rearranging~\eqref{eq:final_term} and by letting $\mathcal{U}^{(T)}$ be the uniform distribution over the queried points $\lbrace \vx_1, \dots, \vx_T \rbrace$ during the run of Algorithm~\ref{alg:iro}, we obtain:
\begin{align*}
    \min_{\secvar \in \secset} \E_{\vx \sim \mathcal{U}^{(T)}} [f(\vx, \secvar)] &\geq \max_{\Pc \in \Delta{(\Xc)}} \min_{\secvar \in \secset} \E_{\vx \sim \Pc}[f(\vx, \secvar)] - \sqrt{\frac{\log(m)}{2T}} - 4\beta_T \sqrt{\frac{\gamma_T \lambda}{T}}.
\end{align*} 
Finally, we require $\epsilon \geq \sqrt{\frac{\log(m)}{2T}} + 4\beta_T \sqrt{\frac{\gamma_T \lambda}{T}}$, which we obtain when $$T \geq \frac{1}{\epsilon^2} \bigg( \frac{\log(m)}{2} + \beta_T \sqrt{32\lambda \gamma_T \log (m)}  + 16 \beta_T^{2} \lambda \gamma_T \bigg).$$
\end{proof}

\section{Proof of Corollary~\ref{corollary}}

\begin{proof} The proof closely follows the one of Theorem~\ref{thm:main}. The main changes are due to the modified best-response rule from~\eqref{eq:modified_best_response}.

For a given distribution $\Qc \in \Delta(\secset)$ and trade-off parameter $\chi \in (0,1]$, we can define the new function
\begin{equation}\label{eq:defn_g}
g(\vx, \secvar) := \chi \cdot f(\vx, \secvar) +(1-\chi) \cdot \E_{\secvar \sim \Qc} [f(\vx, \secvar)]
\end{equation}

Same as before, our goal is to upper bound the difference:
\begin{equation} \label{eq:regret_proof2}
\max_{\Pc \in \Delta{(\Xc)}} \min_{\secvar \in \secset} \E_{\vx \sim \Pc}[g(\vx, \secvar)] - \min_{\secvar \in \secset} \frac{1}{T} \sum_{t=1}^{T} g(\vx_t, \secvar),
\end{equation}
where $\vx_t$ is the point selected at time $t$ by \ouralg using the modified best-response rule as in~\eqref{eq:modified_best_response}.

Next, we condition on the event in Lemma~\ref{conf_lemma} holding true, and we provide upper and lower bounds of the first and second term, respectively.

First, we show that the second term of \eqref{eq:regret_proof2} can be lower bounded as:
\begin{equation} \label{eq:second_term_cor}
    \min_{\secvar \in \secset}  \frac{1}{T} \sum_{t=1}^{T} g(\vx_t, \secvar) \geq  
    \chi \bigg(\min_{\secvar \in \secset} \frac{1}{T} \sum_{t=1}^{T} \oucb_{t-1}(\vx_t,\secvar)\bigg) + (1-\chi)  \bigg(\frac{1}{T} \sum_{t=1}^{T} \mathop{\E}_{\secvar \sim \Qc} [ \oucb_{t-1}(\vx_t,\secvar)]\bigg)  - 4\beta_T \sqrt{\frac{\lambda \gamma_T}{T}} \,.
\end{equation}

To prove Eq.~\eqref{eq:second_term_cor} we make use of \eqref{eq:defn_g} and similar arguments as the ones used in the proof of Theorem~\ref{thm:main}:
\begin{align*}
    & \min_{\secvar \in \secset}  \frac{1}{T} \sum_{t=1}^{T} g(\vx_t, \secvar) = \chi \bigg(  \min_{\secvar \in \secset}  \frac{1}{T} \sum_{t=1}^{T} f(\vx_t, \secvar) \bigg)  + (1-\chi)\bigg( \frac{1}{T} \sum_{t=1}^{T} \mathop{\E}_{\secvar \sim \Qc} [f(\vx_t, \secvar)] \bigg) \\
    & \geq \chi \bigg[ \bigg(\min_{\secvar \in \secset} \frac{1}{T} \sum_{t=1}^{T} \oucb_{t-1}(\vx_t,\secvar)\bigg) -  \frac{2\beta_T}{T} \sum_{t=1}^{T} \sigma_{t-1}(\vx_t, \secvar_t) \bigg] + (1-\chi)\bigg( \frac{1}{T} \sum_{t=1}^{T} \mathop{\E}_{\secvar \sim \Qc} [f(\vx_t, \secvar)] \bigg) \\
        & \geq \chi \bigg[ \bigg(\min_{\secvar \in \secset} \frac{1}{T} \sum_{t=1}^{T} \oucb_{t-1}(\vx_t,\secvar)\bigg) -   \frac{2\beta_T}{T} \sum_{t=1}^{T} \sigma_{t-1}(\vx_t, \secvar_t) \bigg] + (1-\chi)\bigg( \frac{1}{T} \sum_{t=1}^{T} \mathop{\E}_{\secvar \sim \Qc} [\oucb_{t-1}(\vx_t, \secvar) - 2\beta_t\sigma_{t-1}(\vx_t, \secvar)] \bigg)  \\ 
  &  \geq \chi \bigg(\min_{\secvar \in \secset} \frac{1}{T} \sum_{t=1}^{T} \oucb_{t-1}(\vx_t,\secvar)\bigg) + (1-\chi)\bigg( \frac{1}{T} \sum_{t=1}^{T} \mathop{\E}_{\secvar \sim \Qc} [\oucb_{t-1}(\vx_t, \secvar)] \bigg)     - \frac{2\beta_T}{T} \sum_{t=1}^{T} \sigma_{t-1}(\vx_t, \secvar_t)  \\ 
  & \geq \chi \bigg(\min_{\secvar \in \secset} \frac{1}{T} \sum_{t=1}^{T} \oucb_{t-1}(\vx_t,\secvar)\bigg) + (1-\chi)\bigg( \frac{1}{T} \sum_{t=1}^{T} \mathop{\E}_{\secvar \sim \Qc} [\oucb_{t-1}(\vx_t, \secvar)] \bigg) - 4\beta_T \sqrt{\frac{\gamma_T \lambda}{T}}.
\end{align*}

Next, we show that the first term of \eqref{eq:regret_proof2} can be upper bounded as:
\begin{equation}\label{eq:first_term_cor}
    \max_{\Pc \in \Delta{(\Xc)}} \min_{\secvar \in \secset} \E_{\vx \sim \Pc}[g(\vx, \secvar)] \leq \chi \bigg( \frac{1}{T} \sum_{t=1}^T \mathop{\E}_{\secvar\sim \vw_t}[\oucb_{t-1}(\vx_t, \secvar)]\bigg) + (1-\chi) \bigg( \frac{1}{T} \sum_{t=1}^T \mathop{\E}_{\secvar\sim \Qc}[\oucb_{t-1}(\vx_t, \secvar)]\bigg) .
\end{equation}
To prove this we use similar arguments as in the proof of Theorem~\ref{thm:main}:
\begin{align}
     \max_{\Pc \in \Delta{(\Xc)}} \min_{\secvar \in \secset} \E_{\vx \sim \Pc}[g(\vx, \secvar)] 
    & \leq  \max_{\Pc \in \Delta{(\Xc)}}\frac{1}{T} \sum_{t=1}^T  \sum_{i=1}^{m} \vw_t[i] \cdot \E_{\vx \sim \Pc} [g(\vx, \secvar_i)] \nonumber\\ 
    & \leq  \frac{1}{T} \sum_{t=1}^T \max_{\Pc \in \Delta{(\Xc)}} \E_{\vx \sim \Pc} \bigg[ \sum_{i=1}^{m} \vw_t[i] \cdot g(\vx, \secvar_i)] \bigg] \nonumber  \\ 
    & = \frac{1}{T} \sum_{t=1}^T \max_{\vx \in \Xc} \sum_{i=1}^{m} \vw_t[i] \cdot g(\vx, \secvar_i) \nonumber\\ 
    & =  \frac{1}{T} \sum_{t=1}^T \max_{\vx \in \Xc} \bigg[  \chi \cdot \sum_{i=1}^m \vw_t[i] \cdot f(\vx,\secvar_i) + (1- \chi) \cdot \mathop{\E}_{\secvar \sim \Qc} [f(\vx,\secvar)]  \bigg] \nonumber \\
                & \leq  \frac{1}{T} \sum_{t=1}^T \max_{\vx \in \Xc}  \bigg[  \chi \cdot \sum_{i=1}^m \vw_t[i] \cdot \oucb_{t-1}(\vx,\secvar_i) + (1- \chi) \cdot \mathop{\E}_{\secvar \sim \Qc} [\oucb_{t-1}(\vx,\secvar)]  \bigg] \nonumber \\
    &=\chi  \bigg( \frac{1}{T} \sum_{t=1}^{T} \sum_{i=1}^m \vw_t[i] \cdot \oucb_{t-1}(\vx_t,\secvar)\bigg) + (1-\chi)  \bigg(\frac{1}{T} \sum_{t=1}^{T} \mathop{\E}_{\secvar \sim \Qc} [ \oucb_{t-1}(\vx_t,\secvar)]\bigg), \label{eq:first_term_cor6}
    \end{align}
where \eqref{eq:first_term_cor6} is obtained by the rule in \eqref{eq:modified_best_response} used to select $\vx_t$.


Next, we bound the difference in \eqref{eq:regret_proof2} by combining the bounds \eqref{eq:second_term_cor} and \eqref{eq:first_term_cor} and applying~\eqref{eq:MWU_regret_bound} to obtain:
\begin{equation}
    \max_{\Pc \in \Delta{(\Xc)}} \min_{\secvar \in \secset} \E_{\vx \sim \Pc}[g(\vx, \secvar)] - \min_{\secvar \in \secset} \frac{1}{T} \sum_{t=1}^{T} g(\vx_t, \secvar) \leq
    \chi \sqrt{\frac{\log(m)}{2T}}  + 4\beta_T \sqrt{\frac{\gamma_T \lambda}{T}} \label{eq:final_term_cor},
\end{equation}

By letting $\mathcal{U}^{(T)}$ be the uniform distribution over the queried points $\lbrace \vx_1, \ldots, \vx_T \rbrace$ and by using the definitions of $W(\cdot)$ and $\Pc^*$ together with the bound \eqref{eq:final_term_cor}, we obtain:
\begin{align}
  W(\mathcal{U}^{(T)})  = \min_{\secvar \in \secset} \frac{1}{T} \sum_{t=1}^{T} g(\vx_t, \secvar) &\geq \max_{\Pc \in \Delta{(\Xc)}} \min_{\secvar \in \secset} \E_{\vx \sim \Pc}[g(\vx, \secvar)]  -\chi \sqrt{\frac{\log(m)}{2T}}  - 4\beta_T \sqrt{\frac{\gamma_T \lambda}{T}} \nonumber \\ 
  & =  W(\Pc^*) -\chi \sqrt{\frac{\log(m)}{2T}}  - 4\beta_T \sqrt{\frac{\gamma_T \lambda}{T}}
\end{align}

Finally, we require $\epsilon \geq \chi \sqrt{\frac{\log(m)}{2T}} + 4\beta_T \sqrt{\frac{\gamma_T \lambda}{T}}$, which we obtain when $$T \geq \frac{1}{\epsilon^2} \bigg( \frac{ \chi^2 \log(m)}{2} + \chi \:  \beta_T \sqrt{32\lambda \gamma_T \log (m)}  + 16 \beta_T^{2} \lambda \gamma_T \bigg).$$

\end{proof}

\end{document}